\pdfoutput=1

\documentclass[11pt]{article}

\usepackage{natbib}  
\usepackage{url}     

\usepackage[preprint]{acl}

\usepackage{arabtex}
\usepackage{utf8}
\setcode{utf8}

\usepackage{listings}  

\usepackage{times}
\usepackage{latexsym}

\usepackage[T1]{fontenc}

\usepackage[a4paper,margin=2.5cm,heightrounded=true]{geometry}  
\usepackage{graphicx}                       

\usepackage[utf8]{inputenc}

\usepackage{microtype}

\usepackage{inconsolata}

\usepackage{graphicx}

\title{Instructions for *ACL Proceedings}

\author{Passant Elchafei \\
 Ulm University, Germany\\
  \texttt{passant.elchafei@uni-ulm.de} \\\And
  Mervet Abu-Elkheir \\
  German University in Cairo, Egypt \\
  \texttt{mervat.abuelkheir@guc.edu.eg} \\}
  
\begin{document}
\title{Hallucination Detectives at SemEval-2025 Task 3: Span-Level Hallucination Detection for LLM-Generated Answers}
\maketitle

\begin{abstract}
Detecting spans of hallucination in LLM-generated answers is crucial for improving factual consistency. This paper presents a span-level hallucination detection framework for the SemEval-2025 Shared Task, focusing on English and Arabic texts. Our approach integrates Semantic Role Labeling (SRL) to decompose the answer into atomic roles, which are then compared with a retrieved reference context obtained via question-based LLM prompting. Using a DeBERTa-based textual entailment model, we evaluate each role’s semantic alignment with the retrieved context. The entailment scores are further refined through token-level confidence measures derived from output logits, and the combined scores are used to detect hallucinated spans. Experiments on the Mu-SHROOM dataset demonstrate competitive performance. Additionally, hallucinated spans have been verified through fact-checking by prompting GPT-4 and LLaMA. Our findings contribute to improving hallucination detection in LLM-generated responses.
\end{list}
\end{abstract}

\section{Introduction}
LLMs have demonstrated remarkable capabilities in generating human-like text, enabling a wide range of applications, including question-answering, summarization, and conversational agents. However, these models often produce hallucinations that appear plausible but are factually incorrect or unsupported by the input context \citep{quevedo2024detecting}.

Current research efforts focus on different approaches to detect and mitigate hallucinations. Some methods rely on sentence-level classification, where the entire response is labeled factual or not \cite{fadeeva2024fact}. While these techniques provide a general assessment, they lack the granularity to pinpoint the specific hallucinated segments, leading to challenges in localized error correction \cite{liu2021token}. Other works explore token-level approaches, which utilize features such as minimum and average token probabilities to identify hallucinated content \cite{luo2024hallucination}. This strategy shows promising results, but may struggle to effectively combine probabilistic and semantic information.
These issues become more pronounced in morphologically rich and linguistically diverse languages such as Arabic, where complex syntactic rules and dialectal variations add layers of difficulty \cite{wang2023survey}. Recent research highlights the need for improved hallucination detection techniques tailored to Arabic and other low-resource languages \cite{mubarak2024halwasa}.

Our research introduces a span-level hallucination detection framework to address these challenges. Unlike previous sentence-level or token-level approaches, our framework identifies hallucinated spans by decomposing LLM-generated answers into semantic units using Semantic Role Labeling (SRL) and dependency parsing. These units are evaluated against context retrieved using GPT-4, then contradiction detection using pretrained textual entailment BERT model. Simultaneously, token-level confidence scores are computed to capture the model's certainty for each unit. The combined scores provide a refined measure of factuality which indicates hallucination parts.
We evaluate our framework for both English and Arabic languages using the multilingual Mu-SHROOM dataset as part of the SemEval-2025 shared task. To further strengthen reliability, we incorporate an LLM-based verification step by using fact-checking technique using 2 different LLMs (GPT-4, and LLaMA).

The remainder of this paper is organized as follows: Section 2 reviews related work. Section 3 introduces the dataset's structure. Section 4 presents our methodology. Section 5 outlines the experimental results. Finally, Section 6 concludes with key findings and future work.
\section{Related Work}
Hallucination detection in LLMs has been studied, with research mainly focusing on sentence-level classification. These approaches determine whether an entire response is factual or hallucinated, but lack the granularity to identify specific hallucinated spans. Token-based methods leverage confidence measures to estimate factuality, but often fail to account for semantic inconsistencies within generated responses \cite{wang2023survey}. Recent advances have emphasized the need for hallucination detection at the spectral level, allowing a finer-grained factual assessment \cite{ji2023survey}.

Reference-based methods compare generated text against external sources such as Factcheck-Bench \cite{qiu2023detecting}, a fine-grained fact-checking benchmark, and HALoGen \cite{ravichanderhalogen}, a large-scale hallucination evaluation suite that categorizes hallucination errors. Other techniques, such as InterrogateLLM \cite{varshneystitch}, employ self-consistency verification, where LLMs are prompted multiple times to detect contradictions in their responses. Although these methods improve factual verification, they operate primarily at the response level rather than identifying hallucinated spans.

Textual entailment models have also been explored for hallucination detection, classifying responses into entailment, contradiction, or neutrality \cite{wadden2022scifact}. However, these approaches are typically sentence-level, limiting their effectiveness for pinpointing hallucinated spans\cite{chen2023phoenix}.
\begin{figure*}
  \centering
  \includegraphics[width=0.9\textwidth,height=5cm]{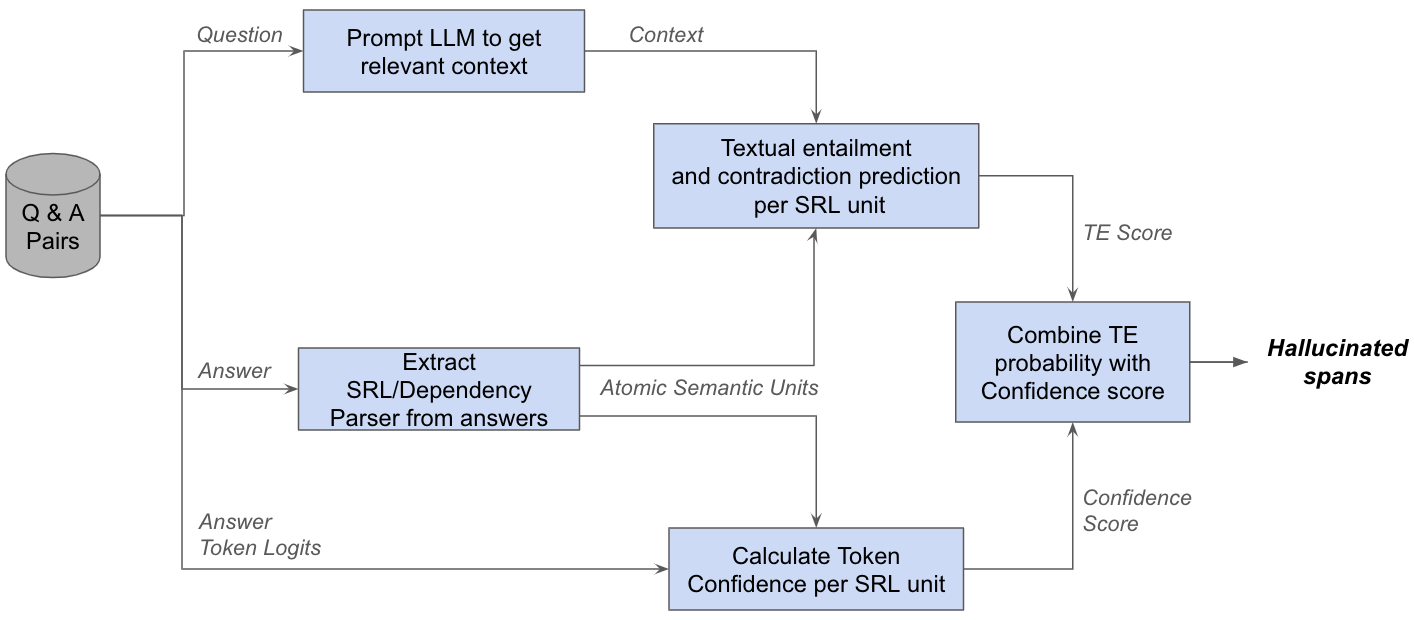}
  \caption{Span-Level Hallucination Detection Framework}
  \label{fig:framework}
\end{figure*}
The detection of low resources languages hallucinations presents additional challenges, particularly in morphologically rich languages like Arabic, where syntactic complexity and dialectal variations complicate the factual verification \cite{10820541}. Datasets such as Halwasa \cite{mubarak2024halwasa} and ACQAD \cite{sidhoum2022acqad} were developed to help in analyzing factual and linguistic inaccuracies. Research on hallucination detection in Arabic has largely focused on sentence-level classification, leaving a gap in span-level hallucination identification \cite{abdelazim2024multi}. Some studies have introduced dependency parsing techniques and Semantic Role Labeling (SRL) to improve hallucination detection which highlight the role of syntactic decomposition in improving the evaluation of facts \cite{liu2023mitigating}.

Advances in SRL for information extraction should also contributed to hallucination detection, particularly in low-resource language. Unlike traditional information extraction tasks, SRL-based techniques focus on verifying atomic claims within generated text, enhancing factual alignment with external sources. Although recent studies focus on hallucination detection, there remains a need for more linguistically adaptive approaches to improve the factual consistency in LLM-generated text.
\section{Dataset Structure}
The dataset used in this research is provided by the SemEval-2025 shared task, Mu-SHROOM (Multilingual Shared-task on Hallucinations and Related Observable Over generation Mistakes). This dataset contains examples in multiple languages, including English, Arabic, Spanish, and French. Each example in the dataset comprises a question-answer pair, the corresponding LLM-generated output, and annotations for hallucinated spans.
\subsection{Dataset Schema}
Each data entry in the dataset includes the following fields:
\begin{itemize}
    \setlength{\itemsep}{0pt}  
    \setlength{\parskip}{0pt}  
    \setlength{\parsep}{0pt}   
\item ID: Unique identifier (e.g. "val-en-1").
\item Language: Question and answer language ("EN" for English, “AR” for Arabic).
 \item Question: Question provided to the model (e.g. "What did Petra van Staveren win a gold medal for?", “{\small\<كم مرحلة يتكون منها رجيم دوكان؟ >}”).
 \item Answer: LLM-generated answer (e.g. "Petra van Stoveren won a silver medal in the 2008 Summer Olympics in Beijing, China.", {\small"\<يعتمد عدد المراحل في رجِيم دُوْكَان على النسخة المحددة >"}
 \newline
{\small"\<من البرنامج، لكن عادةً ما يتألف من خمسة مراحل.>"}
 \item Model Information: Model that generated the output (e.g., "tiiuae/falcon-7b-instruct", "openchat/openchat-3.5-0106-gemma").
 \item Soft Labels: Span-level annotations indicating parts of the text that may be hallucinated, with associated probabilities, e.g. {"start": 10, "prob": 0.2, "end": 12}.
 \item Hard Labels: Annotated spans confirmed as hallucinations, represented by fixed token positions (e.g. [25, 31], [45, 49]).
 \item Model Output Tokens: Tokens of the answer.
 \item Model Output Logits: List of logits (one per token), reflecting the confidence of the model for the prediction of each token. This is the field that is used to calculate the confidence score in our approach.
\end{itemize}
\section{Span-level Hallucination Detection Methodology}
Our framework detects hallucinated spans in LLM-generated answers by combining semantic analysis with token-level probabilistic measures, as described in Figure~\ref{fig:framework}. The system comprises several components: context retrieval for the question, answer decomposition, confidence scoring, textual entailment, score integration to get refined score which indicates the hallucination parts. The framework is designed to handle both English and Arabic, utilizing tools and models suitable for each language. 
\subsection{Context Retrieval}
Given an input question, the relevant context is retrieved using the GPT-4 model with a well-written prompt. Separating retrieval from answer generation ensures an independent factual grounding mechanism. Moreover, retrieving relevant context is more reliable and less complex than generating a well-structured answer, as it relies on matching and ranking mechanisms, whereas answer generation requires reasoning, synthesis, and fluency while ensuring factual correctness. Based on this, the retrieved context serves as a factual reference for evaluating the generated answer and is expected to contain key facts necessary to comprehensively address the question. Additionally, considering the nature of the dataset, which consists of general knowledge question-answer pairs rather than specialized domain-specific queries, this approach is highly suitable. For example, for the question "What did Petra van Staveren win a gold medal for?" the retrieved context may include biographical details about her achievements in swimming, ensuring a factual basis for assessment.
\subsection{Answer Decomposition}
\begin{figure*}
  \includegraphics[width=\textwidth,height=4cm]{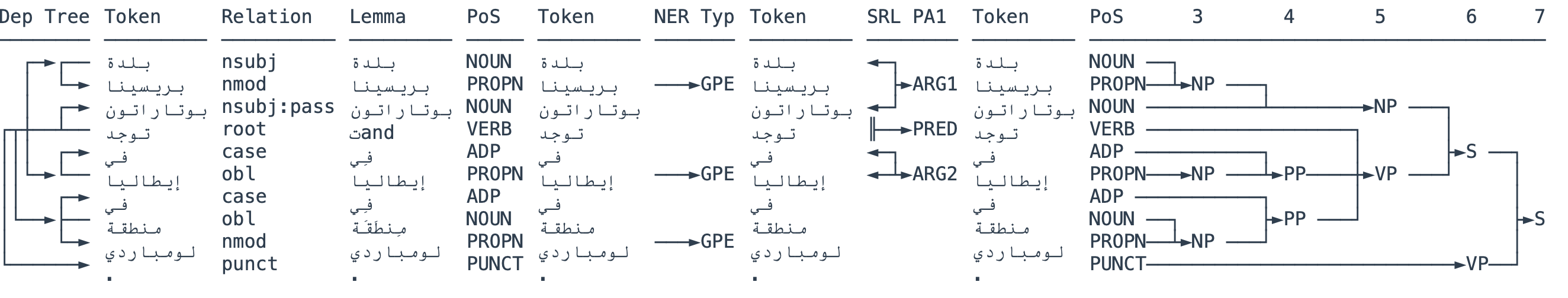}
  \caption{Example of Arabic sentence SRL extraction}
   \label{fig:parsing_1}
\end{figure*}

The generated answer is decomposed into atomic units using Semantic Role Labeling model (SRL). This step extracts structured components (e.g., predicates, arguments) that represent key facts. Each unit is later evaluated for factual alignment with the retrieved context. Example decomposition for "Petra van Staveren won a silver medal in the men's 10 km walk at the 2008 Summer Olympics":
\begin{itemize}
    \setlength{\itemsep}{0pt}  
    \setlength{\parskip}{0pt}  
    \setlength{\parsep}{0pt}   
\item Verb: "won"
\item ARG0: "Petra van Staveren"
\item ARG1: "a silver medal"
\item ARGM-TMP: "at the 2008 Summer Olympics"
\end{itemize}
The AllenNLP BERT-based SRL model \cite{shi2019simple} is used for English. For Arabic, we conduct two experiments: 1) HanLP's multilingual SRL extraction model, which supports Arabic langauge \cite{he-choi-2021-stem}, with an example of an Arabic sentence shown in Figure \ref{fig:parsing_1}. 2) CamelParser \cite{elshabrawy2023camelparser2} for dependency parsing, followed by an SRL extraction algorithm.

CamelParser2.0 is an open-source Python-based Arabic dependency parser designed for the Columbia Arabic Treebank (CATiB) and Universal Dependencies (UD) formalisms. It processes raw text to perform tokenization, part-of-speech tagging, and morphological analysis, enhancing syntactic parsing, which is essential for accurate semantic role decomposition.
\subsection{Confidence Scoring}
Token-level confidence scores are computed using logits from the LLM output. These logits scores are given within the SemEval dataset for each token as described in the Dataset Structure section. A low score indicates reduced confidence, suggesting potential hallucination. The logit score for each atomic unit is computed as described in Equation~\ref{eq:logit_score}.
Where $n$ represents the total number of tokens per unit.
$\mathit{logit}_i$ denotes the logit value of token $i$.
The denominator $\sum_{j} e^{\mathit{logit}_j}$ represents the normalization factor across all tokens, ensuring that the computed probability is within the valid range [0,1].
The final logit score is the average softmax probability of tokens in the generated output.
\begin{equation}
\label{eq:logit_score}
\mathit{logit\_score} = \frac{1}{n} \sum_{i=1}^{n} \frac{e^{\mathit{logit}_i}}{\sum_{j} e^{\mathit{logit}_j}}
\end{equation}
This formulation effectively captures the confidence level, with lower scores indicating higher potential hallucination.
\subsection{Textual Entailment}
To evaluate factual alignment, each atomic unit is compared with the retrieved context, which is described in Section 4.1, using a natural language inference (NLI) model. We choose the DeBERTa \cite{he2021debertav3} entailment model, which predicts whether the context entails, contradicts, or is neutral to the unit. This step generates a set of probabilities corresponding to the entailment, neutral, and contradiction labels. For instance, given the hypothesis "in the 2008 Summer Olympics in Beijing, China", the model output might be: Entailment: 1.1\%, Neutral: 8.7\%, Contradiction: 90.2\%.
\subsection{Score Integration}
The entailment and confidence scores are combined to produce a refined score for each atomic unit. This score determines the likelihood that the unit is factual. A hyperparameter "$\alpha$" controls the weight of the components of entailment and confidence as described in Equation~\ref{eq:refined_score}.
\begin{equation}
\small
\label{eq:refined_score}
\mathit{refined\_score} = \alpha \cdot \mathit{entailment} + (1 - \alpha) \cdot \mathit{confidence}
\end{equation}

\section{Expriments and Results Analysis}
Our experiments focus on two languages (English and Arabic) and assess performance using both intrinsic evaluation metrics and fact-checking verification with LLMs.
\subsection{Evaluation Metrics}
To measure the accuracy and effectiveness of hallucination detection, we evaluate our framework using Intersection over Union (IoU) and Correlation score (Cor) which are the used mertrics by the shared task \cite{vazquez-etal-2025-mu-shroom}. The IoU metric quantifies the overlap between predicted hallucinated spans and the ground truth annotations, where a higher value indicates improved span-level detection accuracy. The Correlation (Cor) metric evaluates the consistency between predicted hallucination probabilities and the ground truth confidence scores, providing insight into the model's reliability in detecting hallucinated spans.

\subsection{Hallucination Identification}
Units flagged with low refined scores are aggregated and reported as hallucinated spans, as shown in Figure \ref{fig:framework}. This enables fine-grained identification of factual inconsistencies within the answer text. Here is an example generated by our framework, showing that it is a “neutral” entailment but because of low logit score based on the output token logit given by the generated answers dataset. 
Refined Score threshold is set at 0.5 to identify hallucinated spans, serving as a balanced decision point. Units with a refined score below this value are classified as hallucinations. 
\newline

\begin{lstlisting}
"ARG1": {
    "hypothesis": "a silver medal",
    "predicted_label": "neutral",
    "entailment_probabilities": {
        "entailment": 0.7,
        "neutral": 75.3,
        "contradiction": 23.9
    },
    "logit_score": 0.3333333333333333,
    "refined_score": 0.1375,
    "hallucinated": true
}
\end{lstlisting}

\begin{lstlisting}
"ARGM-LOC": {
    "hypothesis": "in the 2008 Summer Olympics in Beijing , China",
    "predicted_label": "contradiction",
    "entailment_probabilities": {
        "entailment": 1.1,
        "neutral": 8.7,
        "contradiction": 90.2
    },
    "logit_score": 0.1111111111111111,
    "refined_score": 0.051,
    "hallucinated": true
}
\end{lstlisting}

\subsection{Experimental Results}
\textbf{English Language Results:}
Our model achieves an IoU of 0.358 and a Cor of 0.322. To further validate hallucinated spans, we use GPT-4 and LLaMA to verify detected hallucinations. GPT-4 matched 83\% of the hallucinated spans, while LLaMA identified 72\%, demonstrating the effectiveness of LLM-based verification. GPT-4 achieved higher verification accuracy, successfully confirming or refuting hallucinated spans more consistently.
\newline
\textbf{Arabic Language Results:}
The performance of our hallucination detection framework in Arabic was evaluated using two different models: 1) HanLP Multilingual model for SRL extraction and 2) CamelParser2.0 dependency parser model followed by SRL extraction algorithm. The HanLP model achieves an IoU of 0.205 and a Cor of 0.159, demonstrating lower performance compared to English. This discrepancy is attributed to the morphological complexity and syntactic variation in Arabic, which make span-level hallucination detection more challenging.
An improvement is observed when employing CamelParser followed by SRL extraction algirthm, yielding an IoU of 0.28 and a Cor of 0.21. The increased accuracy suggests that syntactic parsing before semantic role decomposition provides better structured representations for hallucination detection. Additionally, the results indicate that this approach is more robust to dialectal variations, making it better suited for handling complex Arabic linguistic structures. Overall, the findings confirm that integrating dependency parsing improves hallucination detection in morphologically rich languages like Arabic.
For the Arabic Fact-Checking Verification step, GPT-4 identified 58\% of hallucinated spans, showing comparatively lower performance than in English. This is partly due to the GPT-4 model itself being less accurate with Arabic, particularly when handling ambiguous factual claims. However, we utilized GPT-4 to ensure a consistent evaluation approach between Arabic and English cases.
\section{Conclusion}
This paper presents a span-level hallucination detection framework that integrates Semantic Role Labeling (SRL), textual entailment, and token-level confidence scoring to identify hallucinations in LLM-generated answers. Evaluated on the Mu-SHROOM dataset, our approach achieves an IoU of 0.358 and a Cor of 0.322 in English, and an IoU of 0.28 and a Cor of 2.1 in Arabic when using CamelParser combined with the SRL extraction algorithm.
Our results emphasize the effectiveness of dependency parsing before SRL extraction, particularly in Arabic, where linguistic complexity poses additional challenges. Despite these improvements, challenges remain for morphologically rich languages. Future work will focus on enhancing entailment models and addressing additional syntactic structures, such as nominal sentences in Arabic and long complex sentences.
\bibliography{acl_latex_review_1}

\end{document}